\pdfoutput=1

\documentclass[11pt]{article}

\usepackage[]{emnlp2021}

\usepackage{times}
\usepackage{latexsym}
\usepackage{booktabs}
\usepackage{tabularx}
\usepackage{subfiles}
\usepackage{comment}

\usepackage{scalerel}[2016/12/29]
\newcommand\smallrightarrow{\scaleobj{0.75}{\rightarrow}}

\usepackage{multirow}

\usepackage[normalem]{ulem}

\usepackage[T1]{fontenc}

\usepackage[utf8]{inputenc}
\usepackage{lipsum}
\usepackage{microtype}
\usepackage{todonotes}
\usepackage{enumitem}
\setlist{noitemsep} 

\title{{E}din{S}aar@{WMT}21: North-Germanic Low-Resource Multilingual {NMT}}

\author{Svetlana Tchistiakova\textsuperscript{{\normalfont 1}}{\normalfont,} Jesujoba O. Alabi\textsuperscript{{\normalfont 2}}{\normalfont,} \\ {\bf Koel Dutta Chowdhury}\textsuperscript{{\normalfont 2}}, {\bf Sourav Dutta}\textsuperscript{{\normalfont 3}}, {\bf Dana Ruiter}\textsuperscript{{\normalfont 2}}\\
\textsuperscript{1}The University of Edinburgh, Edinburgh, Scotland \\
\textsuperscript{2}Saarland University, Saarbrücken, Germany \\
\textsuperscript{3}Technical University of Kaiserslautern, Kaiserslautern, Germany \\
Corresponding author: \texttt{stchisti@ed.ac.uk} 
}

\begin{document}
\maketitle
\begin{abstract}

We describe the EdinSaar submission to the shared task of Multilingual Low-Resource Translation for North Germanic Languages at the Sixth Conference on Machine Translation (WMT2021). We submit multilingual translation models for translations to/from Icelandic (\texttt{is}), Norwegian-Bokm\r{a}l (\texttt{nb}), and Swedish (\texttt{sv}). We employ various experimental approaches, including multilingual pre-training, back-translation, fine-tuning, and ensembling. In most translation directions, our models outperform other submitted systems.

\end{abstract}

\section{Introduction}

This paper presents the neural machine translation (NMT) systems jointly submitted by The University of Edinburgh and Saarland University to the WMT2021 Multilingual Low-Resource Translation for Indo-European Languages task, describing both primary and contrastive systems which translate to/from the three North Germanic languages, Icelandic (\texttt{is}), Norwegian-Bokm\r{a}l (\texttt{nb}), and Swedish (\texttt{sv}). Our contrastive system, submitted as ``edinsaarContrastive'' outperforms the other submissions across all evaluation metrics except for BLEU, for which our ``edinsaarPrimary'' system performs best.


Although low-resource MT has recently gained much attention, there is little prior work on North Germanic languages. We contribute to this space by experimenting with both training a multilingual system from scratch and exploiting model adaptation from a large pre-trained language model. We fine-tune our initial translation models to the target languages, and then experiment with further in-domain fine-tuning. Data is sourced from openly available data sets in accordance with the corpora allowed in the shared task. We use parallel data sets pairing our target languages with each other and with the allowed high-resource languages, and monolingual data from Wikipedia. 

The rest of the paper is structured as follows: we review related work in Section \ref{sec:related}, we introduce the methods and experimental settings including data and model architecture in Section \ref{sec:method}, we evaluate model performance in Section \ref{sec:evaluation}, and, finally, we draw conclusions and suggest avenues for future work in Section \ref{sec:conclusion}.

\section{Related Work}\label{sec:related}

Recent work in NMT for North Germanic languages is limited; however, OPUS-MT \cite{tiedemann2020opus}, which contains over 1,000 pre-trained, ready-to-use neural MT models including models for Danish, Norwegian, and Swedish, is a notable exception. 

Due to the scarcity of parallel data for low-resource languages, recent work leverages monolingual data, including pivoting from high-resource languages \cite{currey-heafield-2019-zero,kim2019pivot}, and using back-translation \cite{backtranslation,edunov2018understanding} to generate pseudo-parallel data with synthetic sources from monolingual data. Since the little parallel data that is available often comes from noisy web crawls, parallel corpus filtering is used to develop better translation models \cite{koehn-etal-2020-findings}. Additional methods for boosting the performance of low-resource pairs include transfer learning from models trained on higher-resource pairs \cite{zoph2016transfer, kocmi2018trivial}, and developing multilingual systems to allow models to take advantage of linguistic relatedness. Multilingual systems can employ either separate encoders or decoders for each language \cite{dong-etal-2015-multi,firat-etal-2016-multi}, or shared encoders/decoders, and can additionally make zero-shot MT possible  \cite{johnson-etal-2017-googles,thanhle2016towardmultilingual}, while scaling to hundreds of language pairs \cite{aharoni2019massively,fan2020beyond}. Sampling language pairs in proportion to their prevalence in the training data can ensure that all directions get enough coverage by the model \cite{arivazhagen2019massively,fan2020beyond}. Further fine-tuning multilingual systems on target language directions can improve performance of low-resource pairs \cite{neubig2018rapidadaptation,lakew2019adapting}. Adapting a multilingual pre-trained language model to the translation task has led to improvements in translation quality \cite{clinchant-etal-2019-use,chen-etal-2020-distilling}. Finally, combining multiple MT system checkpoints together by ensembling improves performance of the final system \cite{sennrich-etal-2017-university}.

\section{Method}\label{sec:method}

Given a set of primary languages $L_p$ and secondary languages $L_s$, we train a multilingual MT system on the parallel data between all the language combinations $\{L_p, L_s\} \leftrightarrow \{L_p, L_s\}$. This is our \textbf{baseline}. We extend this approach with a combination of the following methods:

\begin{itemize}
    \item[] \textbf{Pre-training}: We initialize a base model using a highly multilingual pre-trained model, in order to transfer the learned parameters to the translation task. This is our \textbf{primary} system.
    \item[] \textbf{Back-translation}: We use the baseline model to back-translate monolingual corpora in $L_p$ into all other languages in $L_p$ to obtain a training data set of back-translations $D_{\rm BT}$.
    \item[] \textbf{Fine-tuning}: We fine-tune the baseline model on the subset of languages $\{L_p, L_s\} \leftrightarrow L_p$, on both parallel and back-translated data $D_{\rm BT}$. Our \textbf{contrastive} system is an ensemble of the last four checkpoints of this model.
\end{itemize}

\subsection{Data}\label{sec:data}

For training our models, we include data from the target primary low-resource languages, Icelandic (\texttt{is}), Norwegian-Bokm\r{a}l (\texttt{nb}), and Swedish (\texttt{sv}), and the related secondary languages Danish (\texttt{da}), German (\texttt{de}), English (\texttt{en}). 

We use data for all translation directions involving \texttt{da}, \texttt{de}, \texttt{en}, \texttt{is}, \texttt{nb}, \texttt{sv} from the following \textbf{parallel} corpora from Opus: Bible \cite{christodouloupoulos_steedman_2014}, Books \cite{tiedemann-2012-parallel}, Europarl \citep{koehn2005epc}, GlobalVoices \cite{tiedemann-2012-parallel}, JW300 \citep{agic-vulic-2019-jw300}, MultiCCAligned \citep{elkishky_ccaligned_2020}, Paracrawl \cite{espla-etal-2019-paracrawl}, TED2020 \citep{reimers-2020-multilingual-sentence-bert}, and WikiMatrix \citep{WikiMatrix}. We also use all corpora from ELRC\footnote{\url{https://elrc-share.eu/}} that include these directions (a total of 159 corpora, retrieved in May 2021). 
These corpora include all corpora allowed by the shared task, with the exception of the Opus-100 data set, which we avoided as it had many duplicate sentences with the above corpora.

We use \textbf{monolingual} data from Wikipedia for \texttt{is} and \texttt{nb} to augment our data set with back-translations \citep{backtranslation}. Because the Wikipedia data for \texttt{sv} was created in large part by a bot\footnote{\url{https://en.wikipedia.org/wiki/Lsjbot}} and consisted of many stub articles and tables, we use the \texttt{sv} portion of our training data as monolingual data for back-translation instead.

Our final data includes 30 language directions:



\begin{enumerate}[label=(\alph*)]
\item $L_p \leftrightarrow L_p$:  \texttt{\{is,nb,sv\}} $\leftrightarrow$ \texttt{\{is,nb,sv\}}
\item $L_p \leftrightarrow L_s$:   \texttt{\{is,nb,sv\}} $\leftrightarrow$  \texttt{\{da,de,en\}}
\item $L_s \leftrightarrow L_s$: \texttt{\{da,de,en\}} $\leftrightarrow$ \texttt{\{da,de,en\}}
\item $L_{p\_bt}\ \smallrightarrow L_p$: \texttt{\{is,nb,sv\}} $\smallrightarrow$ \texttt{\{is,nb,sv\}}
\end{enumerate}

where $L_{p\_bt}$ is created from the monolingual target side back-translated data $D_{\rm BT}$.

\begin{table*}
\centering
\small
\begin{tabular}{|l|rl|rl|rl|rl|rl|rl|}
\hline
            & \textbf{de} & & \textbf{en} & & \textbf{is} & & \textbf{nb} & & \textbf{sv} & \\ \hline
\textbf{da}  & 6921831 & (48) & 20604309 & (77) & 797806  & (68) & 10654 & (89) & 5590356    & (65) \\ \hline
\textbf{de}  &         &     & 144890166 & (80) & 456054  & (62) & 24963 & (91) & 5119372    & (59) \\ \hline
\textbf{en}  &         &     &           &      & 3766342 & (78) & 279370  & (46) & 21906032 & (78) \\ \hline
\textbf{is}  &         &     &           &      & 351833  & (60) & 597     & (89) & 446106   & (46) \\ \hline
\textbf{nb}  &         &     &           &      &         &      & 2943733 & (44) & 14247    & (89) \\ \hline
\end{tabular}
\caption{Number of sentences after filtering (with \% of total raw data remaining after filtering) in each language direction from source (left) to target (top) from all corpora and for additional monolingual data from Wikipedia. The parallel data was mirrored in the reverse directions to create 30 total language directions for training.}
\label{tab:filtered-counts}
\end{table*}

\paragraph{Parallel Data Filtering} \label{sec:parallel-filtering}

We filter the parallel data using rule-based heuristics borrowed from the Bifixer/Bicleaner tools \citep{prompsit:2018:WMT, prompsit:2020:EAMT} and language identification using FastText \cite{joulin2016fasttext, joulin2016bag}. This repairs common orthographic errors, including fixing failed renderings of glyphs due to encoding errors, replacing characters from the wrong alphabet with correct ones, and un-escaping html. It also removes any translation pairs where: the pair is a duplicate, the source and target are identical, the source or target language is not the intended language, one side is more than $2$x the length of the other, one side is empty, one side is longer than $5000$ characters, one side is shorter than 3 words, or one side contains primarily URLs and symbols rather than text.

Filtering reduces our parallel data to $77\%$ of its original total size. This data is then reversed in order to train our multilingual model in all translation directions, resulting in a total of 421,656,410 parallel sentence pairs in all 30 language directions. Table \ref{tab:filtered-counts} lists the filtered data counts and the percentage of the original data that these counts represent.

\paragraph{Monolingual In-Domain Data Filtering}\label{sec:monolingual-filtering}

The validation set provided by the shared task organizers, containing thesis abstracts and descriptions, is dissimilar to our available parallel corpora. Therefore, we filter the Wikipedia monolingual \texttt{is} and \texttt{nb} data for similarity to this validation set to create in-domain monolingual data for use in back-translation. We identify in-domain monolingual instances in our data by calculating the cosine similarity between each sentence in a given language in the monolingual data to each of the sentences in the shared task validation data for that language. When a training instance has a similarity of $>=\theta$ with at least one validation instance, it is added to the in-domain fine-tuning corpus. We set $\theta=0.9$ and use LASER
\citep{artetxe2018massively} to extract vector representations of sentences for calculating similarity.


\paragraph{Validation and Test Data}
\label{sec:valid}

We split off 2000 sentence pairs from each language pair in our parallel data to use as an \textbf{internal test set}.
For \texttt{is-nb} directions, we use the few parallel sentences available for this, meaning that no parallel data is left for the training or validation corpus. Therefore, translating between these directions is a zero-shot task for our models. 


We also split off 2000 sentence pairs from each language pair in our parallel data for \textbf{internal validation}.  For validation of our primary model, we use the entire collection of 2000 validation sentence pairs in each language direction. For the baseline system, we cut this down to a total of $\sim2000$ sentences, because performing validation is quicker on smaller data. Therefore, we use a subset of 72 validation sentences in each $\{L_p,L_s\} \leftrightarrow \{L_p,L_s\}$, except \texttt{is-nb}, resulting in 2016 sentences. For the contrastive model, we use the same sentences in only $\{L_p,L_s\} \leftrightarrow \{L_p\}$, to which we add 72 sentences from the back-translated data in the \texttt{is-nb} directions, resulting in a total of 1728 sentences.

We use the \textbf{shared task validation} set, to compare performance between our systems, and do not use it during model training or fine-tuning. We additionally report results Section \ref{sec:evaluation} on the \textbf{shared task test set}, which was provided to the teams after the completion of the shared task. These test sets contain approximately 500 sentences in each language direction.

\paragraph{Back-translation}\label{sec:backtranslation}

We use the baseline system (Section \ref{sec:baseline}) to create back-translations of our monolingual in-domain filtered Wikipedia data. This generates synthetic sources from \texttt{is} to \{\texttt{nb}, \texttt{sv}\} and from \texttt{nb} to \{\texttt{is}, \texttt{sv}\}. We additionally back-translate the \texttt{sv} side of our parallel \texttt{nb-sv} corpus into \texttt{is} and our \texttt{is-sv} corpus into \texttt{nb}. After creating the back-translations, we filter the new synthetic parallel data sets again using the parallel data filtering steps (Section \ref{sec:parallel-filtering}), in order to remove sentences that consisted primarily of model errors or hallucinations. The final counts of filtered back-translated data are in Table \ref{tab:bt}, as well as the percentage of the original total in-domain data that these counts represent.

\begin{table}
\centering
\small
\begin{tabular}{|l|r|r|r|}
\hline
            & \textbf{is}  & \textbf{nb}  & \textbf{sv}           \\ \hline
\textbf{is} &                  & 2564234  (87)    & 10123   (99)  \\ \hline
\textbf{nb} & 279818  (80)     &                  & 344583  (78)  \\ \hline
\textbf{sv} & 299277  (85)     & 2521823  (86)    &               \\ \hline
\end{tabular}
\caption{Number of back-translated filtered sentences (with \% of total data remaining after filtering) between synthetic source (left) to original target (top).}
\label{tab:bt}
\end{table}

\subsection{Byte-pair Encoding}\label{sec:bpe}

To create a vocabulary for our baseline and contrastive systems, we train a shared byte-pair encoding (BPE) \citep{sennrich-etal-2016-neural} model using SentencePiece \citep{SPM}. 
We sample 10 million monolingual sentences from our parallel training data, based on the amount of monolingual data available for each language. Following the idea of \citet{arivazhagen2019massively}, we use temperature sampling, where the probability of sampling any particular data set $D$ in language $\ell$ out of the $n$ total data sets is defined as $p_\ell = (\frac{D_\ell}{\sum_i^n{D_i}})^\frac{1}{T}$, where we set $T=5$. The goal of sampling in this way is to provide a compromise that allows the BPE model to view a larger portion of lower resource language tokens (unlike sampling according to the original distribution would), while still providing extra space in the model for the larger variety of tokens coming from high-resource corpora (unlike sampling uniformly would).  We use a vocabulary of $32,000$ tokens. When BPE-ing our training data, we use BPE-dropout \cite{provilkov-etal-2020-bpe} with a probability of $0.1$.

\subsection{Models}\label{sec:models}

\paragraph{Baseline}\label{sec:baseline}

Our baseline system is trained on a concatenation of data sets (a), (b), and (c) (see Section \ref{sec:data}). The data is pre-processed using byte-pair encoding as described in Section \ref{sec:bpe}. Following the method of \citet{johnson-etal-2017-googles}, we jointly train the model to translate in all our language directions, pre-pending a token \texttt{<2xx>} to the source side to inform the model which target language to translate into. The system is comprised of a transformer base model trained using Marian \cite{mariannmt} with cross-entropy loss, following the method of \cite{vaswani} and the default Marian \texttt{transformer} configuration.

We differ from the default configuration in the following ways. We fit our mini-batch to a workspace of $6144$ MB, set the learning rate to $0.0003$ with a warm-up increasing linearly for $16000$ batches and decaying by $\frac{16000}{\sqrt{no.\ batches}}$ afterwards. We train on multiple GPUs using Adam \citep{adam} with synchronous updates for optimization, setting $\beta_1=0.9$, $\beta_2=0.98$ and $\epsilon=1e-09$. We set transformer dropout between layers to $0.01$. We use a maximum sentence length of $200$ tokens, a maximum target length as source length factor of $2$, and a label smoothing of $0.01$. During validation, we use a beam size of $6$ and normalize the translation score by $translation\_length^{0.6}$. We check translation quality on our internal validation set (Section \ref{sec:valid}) every 5000 model updates and stop training when performance doesn't improve for 15 checkpoints. The model was trained for approximately 66 hours on four NVIDIA GeForce RTX 3090 GPUs.

\paragraph{Contrastive}\label{ref:contrastive}

Our contrastive model fine-tunes the baseline model directly, using a concatenation of all data sets that incorporate our target languages, including parallel and back-translated data (the data sets (a), (b), and (d) described in Section \ref{sec:data}). The fine-tuned model uses the same architecture, training settings, and stopping criterion as the original baseline model, essentially allowing us to continue training further from the original baseline. The final submitted system is an ensemble of the last four checkpoints of this model. The model was trained for approximately 54 hours on two NVIDIA GeForce RTX 2080 TI GPUs.

\paragraph{Primary}\label{ref:primary}

For the primary system, we adapt mt5 \cite{mt5},
a multilingual pre-trained transformer language model, to the translation task.
We use mt5 because of its state-of-the-art performance and its coverage of all of our target North Germanic languages.
We use the SimpleTransformers\footnote{\url{https://github.com/ThilinaRajapakse/simpletransformers}} framework which extends  HuggingFace \cite{wolf2019huggingface}, with the default parameters. Since our model is initialized from the parameters of the \texttt{mt5-base} system, including the embedding layers, we use the same byte-pair encoded vocabulary as the original model. Due to resource constraints, we sample a total of $100$k parallel sentences from data sets (a), (b) and (c) (described in Section \ref{sec:data}). We pre-pend a string to the source side to indicate to the model which target language to translate into, and adapt the model for 5 epochs. 
We further fine-tune this model on 100k additional sentences that include our target languages from sets (a) and (b) to create our Primary system. The model was trained for approximately 46 hours on a single NVIDIA A100 SXM4 GPU.

\begin{table*}
\centering
\small
\begin{tabular}{|c|l|r|r|r|r|r|r|r|}
\hline
\textbf{} & \textbf{Model} & \textbf{is $\rightarrow$ nb} & \textbf{is $\rightarrow$ sv} & \textbf{nb $\rightarrow$ is} & \textbf{nb $\rightarrow$ sv} & \textbf{sv $\rightarrow$ is} & \textbf{sv $\rightarrow$ nb} & \textbf{Avg.} \\ \hline

\multirow{5}{*}{\textbf{Internal test}}
& marian             & 12.5   & 33.3  & 11.8   & 26.7  & 27.8  & 18.7 & 21.8 \\
& marian\_ft         & 19.1   & 41.7  & 16.1   & 31.6  & 38.4  & 30.3 & 29.5 \\
& marian\_ft\_esmb   & 19.3   & 42.2  & 16.4   & 31.6  & 39.2  & 30.3 & 29.8 \\
& mt5\_base\_ada     & 23.1   & 42.3  & 19.4   & 33.7  & 42.8  & 33.9 & 32.5 \\
& mt5\_base\_ada\_ft & \textbf{26.5}  & \textbf{42.9} & \textbf{20.0} & \textbf{33.9} & \textbf{43.3} & \textbf{34.2}  & \textbf{33.5} \\ \hline

\multirow{5}{*}{\textbf{Shared valid}} 
& marian             & 10.9    & 13.5  & 15.1  & 41.3  & 12.2 & 24.9  & 19.7 \\ 
& marian\_ft         & 13.0    & 18.0  & 22.9  & 50.0  & 19.4 & 45.9  & 28.2 \\
& marian\_ft\_esmb   & 13.9    & 18.2  & 23.6  & \textbf{50.6}  & 20.1  & \textbf{46.7} & 28.8 \\ 
& mt5\_base\_ada     & 14.6    & \textbf{19.2} & 25.8  & 46.6   & 20.6  & 43.2  & 28.3  \\ 
& mt5\_base\_ada\_ft & \textbf{17.4}  & 18.7  & \textbf{26.5}  & 47.9  & \textbf{20.8}  & 44.2  & \textbf{29.3} \\ \hline

\multirow{2}{*}{\textbf{Shared test}}
& marian\_ft\_esmb       & 13.0  & 17.3  & 18.3  & \textbf{45.4}  & 20.2  & \textbf{48.2} & 27.1 \\ 
& marian\_base\_ada\_ft  & \textbf{16.3}  & \textbf{18.8}  & \textbf{19.5}  & 42.9  & \textbf{22.4}  & 45.4 & \textbf{27.5} \\ \hline

\end{tabular}
\caption{SacreBLEU (detokenized) results on the internal test set and the shared task validation and test sets.}
\label{tab:europeana-results}
\end{table*}

\section{Evaluation}\label{sec:evaluation}
Table \ref{tab:europeana-results} reports results on detokenized SacreBLEU on 
each of our internal test set, the shared task validation set, and the shared task test set\footnote{BLEU+case.mixed+numrefs.1+smooth.exp\linebreak +tok.13a+version.1.4.14}.
Comparing results on the internal test set and shared task validation sets show that our models fail to generalize well to the shared task domain, in particular, in the least represented languages in our data set such as \texttt{is}.
In future work, we would like to experiment with different sampling methods to boost the performance of the least represented directions.



Comparing results between models, our primary \texttt{mt5\_base\_ada} system outperforms the  \texttt{marian} model trained from scratch by an average of $+10.7$ and $+8.6$ BLEU points on the internal and shared task validation sets, respectively. 
The further fine-tuned variant \texttt{mt5\_base\_ada\_ft} leads to an additional average improvement of just under $+1$ BLEU point on both sets, showing that the mt5 model already learned a good amount about our target task and languages from our initial adaptation step. The \texttt{marian} model is also outperformed by 
the fine-tuned variant \texttt{marian\_ft}, resulting in an average improvement of $+7.7$ BLEU points on the internal test set and $+8.5$ BLEU points on the shared task validation set.

Both the \texttt{mt5\_base\_ada\_ft} and \texttt{marian\_ft} models are exposed to similar language data; however, the mt5 language model we adapted from (\texttt{mt5-base}) is much larger than our \texttt{marian} model ($580$ million vs $44$ million
parameters), and was trained on more language data ($750$ GB vs $46$ GB), so it had a much stronger base to start from. Ensembling the last $4$ checkpoints of the fine-tuned marian model for \texttt{marian\_ft\_esmb} boosts performance by $+0.3$ and $+0.6$ average BLEU on the internal and shared task validation sets over \texttt{marian\_ft}; however, the \texttt{mt5\_base\_ada\_ft} model still outperforms the \texttt{marian\_ft\_esmb} model by $+3.7$ and $+0.5$ average BLEU on the internal test set and the shared task validation set, respectively. Therefore, we submitted the \texttt{mt5\_base\_ada\_ft} model as our primary system to the shared task; however, our contrastive system, the \texttt{marian\_ft\_esmb} model, won in the shared task rankings.



In the global automated evaluations of the shared task,
our contrastive system is the best-performing submitted system\footnote{Only our primary model was submitted for manual evaluation, where it outranked the other submissions. Official rankings are available at: \url{http://statmt.org/wmt21/multilingualHeritage-translation-task.html}}, 
outperforming the official mt5 baseline by approximately $+8.5$ BLEU. We hypothesize that the mt5 baseline, while being pre-trained on massive amounts of partially noisy monolingual data, has learned the translation task via training on the development set only, so it has less informative parallel data available than our models. The M2M-100 \cite{fan2020beyond} baseline outperforms all submitted systems, despite having been trained on noisy parallel data only. We hypothesize that the highly-multilingual nature of the M2M-100 model allows the target languages to benefit from the supervisory signals between related language combinations. 

\section{Conclusion and Future Work}\label{sec:conclusion}
We contribute to the growing space of NMT for North Germanic languages. We explore multilingualism by training a transformer with a shared encoder and decoder for all language pairs from scratch, as well as adapting a pre-trained multilingual language model. Fine-tuning these models to our low-resource language pairs was a key component in our success in the task, and we additionally confirm that employing popular techniques in machine translation, such as data filtering, back-translation, and model ensembling are beneficial for improving performance on low-resource directions.
In future work, we would like to experiment with fine-tuning 
additional pre-trained models such as the M2M-100, incorporating iterative back-translation, and trying different sampling methods during training to boost lower performing low-resource language pairs. 

\section*{Acknowledgements}

The authors thank the University of Edinburgh and the German Research Center for Artificial Intelligence (DFKI GmbH) for providing the necessary infrastructure and resources to run the experiments. 

This research is based upon work supported in part by the Office of the Director of National Intelligence (ODNI), Intelligence Advanced Research Projects Activity (IARPA), via contract \#FA8650-17-C-9117. The views and conclusions contained herein are those of the authors and should not be interpreted as necessarily representing the official policies, either expressed or implied, of ODNI, IARPA, or the U.S. Government. The U.S. Government is authorized to reproduce and distribute reprints for governmental purposes notwithstanding any copyright annotation therein.

\bibliography{anthology,custom}
\bibliographystyle{acl_natbib}

\end{document}